\documentclass[fleqn,10pt]{wlscirep}
\usepackage[utf8]{inputenc}
\usepackage[T1]{fontenc}

\usepackage{enumitem}
\usepackage{soul}
\usepackage{xcolor,colortbl}
\usepackage{multirow}
\usepackage{tcolorbox}
\usepackage{graphicx}
\usepackage{float}
\usepackage{subfig}
\usepackage{pifont}
\usepackage{todonotes}

\NewDocumentCommand{\heng}{mO{}}{\textcolor{red}{\textsuperscript{\textit{Heng}}\textsf{\textbf{\small[#1]}}}}

\title{Factuality Challenges in the Era of Large Language Models}

\author[1]{Isabelle Augenstein}
\author[2]{Timothy Baldwin}
\author[3]{Meeyoung Cha}
\author[4]{Tanmoy Chakraborty\footnote{Corresponding author, Email: \url{tanchak@iitd.ac.in}}}
\author[5]{Giovanni Luca Ciampaglia}
\author[6]{David Corney}
\author[7]{Renee DiResta}
\author[8]{Emilio Ferrara}
\author[9]{Scott Hale}
\author[10]{Alon Halevy}
\author[11]{Eduard Hovy}
\author[12]{Heng Ji}
\author[13]{Filippo Menczer}
\author[14]{Ruben Miguez}
\author[2]{Preslav Nakov}
\author[15]{Dietram Scheufele}
\author[4]{Shivam Sharma}
\author[16]{Giovanni Zagni}

\affil[1]{University of Copenhagen, Nørregade 10, 1172 København, Denmark}
\affil[2]{Mohamed bin Zayed University of Artificial Intelligence, Masdar City, Abu Dhabi, 7909, United Arab Emirates}
\affil[3]{Korea Advanced Institute of Science and Technology, 291 Daehak-ro, Yuseong-gu, Daejeon, South Korea}
\affil[4]{Indian Institute of Technology Delhi, New Delhi, 110016, India}
\affil[5]{University of Maryland, College Park, Maryland 20742, USA}
\affil[6]{Full Fact, 17 Oval Way, London, SE11 5RR, United Kingdom}
\affil[7]{Stanford University, 450 Jane Stanford Way, Stanford, CA 94305, USA}
\affil[8]{University of Southern California, Los Angeles, CA 90007, USA}
\affil[9]{University of Oxford, Broad St, Oxford OX1 3AZ, United Kingdom}
\affil[10]{Meta AI, 1 Hacker Way, Menlo Park, CA 94025, USA}
\affil[11]{Carnegie Mellon University, 5000 Forbes Ave, Pittsburgh, PA 15213, USA}
\affil[12]{University of Illinois Urbana-Champaign, 506 S. Wright St. Urbana, IL 61801-3633, USA}
\affil[13]{Indiana University, 1015 E 11th St., Bloomington, IN 47408, USA}
\affil[14]{Newtrales, C/Vandergoten 1, 28014 Madrid, Spain}
\affil[15]{University of Wisconsin, Madison, WI, USA}
\affil[16]{Pagella Politica/Facta, viale Monza 259/265, Milano, 20125, Italy}

\begin{abstract}
The emergence of tools based on Large Language Models (LLMs), such as OpenAI's ChatGPT, Microsoft's Bing Chat, and Google's Bard, has garnered immense public attention. These incredibly useful, natural-sounding tools mark significant advances in natural language generation, yet they exhibit a propensity to generate false, erroneous, or misleading content --- commonly referred to as \emph{hallucinations}. Moreover, LLMs can be exploited for malicious applications, such as generating false but credible-sounding content and profiles at scale. This poses a significant challenge to society in terms of the potential deception of users and the increasing dissemination of inaccurate information. In light of these risks, we explore the kinds of technological innovations, regulatory reforms, and AI literacy initiatives needed from fact-checkers, news organizations, and the broader research and policy communities. By identifying the risks, the imminent threats, and some viable solutions, we seek to shed light on navigating various aspects of veracity in the era of generative AI.
\end{abstract}

\begin{document}

\flushbottom
\maketitle

\thispagestyle{empty}

\section{Introduction}

In his pioneering work on information theory published in 1948\cite{shannon1948mathematical}, Claude Shannon proposed simple statistical language models based on letter and word frequencies capable of generating text that, although devoid of meaning, resembled human language. With the vast amounts of digital text available for training and thanks to advances in computation, large language models (LLMs) today are capable of memorizing and reasoning about natural language\cite{huang-chang-2023-towards}; they can generate text that human readers find increasingly hard to distinguish from actual human text. From early LLMs such as GPT\cite{radford2018improving} to more recent models such as GPT-4\cite{openai2023gpt4} and LLaMA 2\cite{touvron2023llama}, several iterations of these technologies have achieved unprecedented sophistication in natural language understanding and generation ability. Despite the differences in their architectures, training data, and algorithms, LLMs principally aim to address the same task, known as ``next-word prediction'' --- given a sequence of words as a prompt, what word(s) should follow? 

LLMs were brought to the general public's attention in late 2022 with the release of OpenAI's ChatGPT\cite{openaiIntroducingChatGPT}, a chatbot based on LLM technology specifically trained to generate responses as part of a conversation. Analyst reports found that ChatGPT reached 100 million monthly active users only two months after its launch, making it the fastest-growing consumer application in history\cite{hu2023chatgpt}. Subsequent innovations by large tech companies such as Microsoft, Meta, and Google\cite{thoppilan2022lamda} pushed the technology's profile even further. These were then followed by Alpaca\cite{alpaca} and Vicuna\cite{vicuna}, open-source alternatives created by instruction-tuning Meta's LLaMA base model\cite{touvron2023llama} with conversations from ChatGPT. Similarly, Claude\cite{claude}, Falcon\cite{falcon}, Jurassic\cite{jurassic}, and Jais\cite{jais} were developed through instruction-tuning of their own individual LLM models.


The proliferation of LLMs has led to concerns about an arms race in Generative AI (GenAI) among leading tech companies. In 2020, prior to eventually open-sourcing the second release of their \text{generative pre-trained transformer} (GPT) series, GPT-2\cite{radford2019language}, OpenAI cited the potential for societal harm as one of the reasons for not releasing the source code of their first model, GPT\cite{brown2020language}. Since then, the industry has largely abandoned such a cautious stance. 
As a result, LLMs are rapidly gaining popularity among a vast number of professionals and everyday people with limited or no technical knowledge. On the research front, a recent survey indicated a significant uptick in the number of papers containing the term ``large language model'' in their title or abstract~\cite{LLMSurvey}.

The year 2022 brought a dramatic increase in the parameter size of these models, reminiscent of Moore's law, and a qualitative leap in commonsense reasoning~\cite{LLMSurvey} and knowledge generalization~\cite{hierarchicalschema2023}. 
Some studies also point out that LLMs with billion-scale parameters have \emph{emergent abilities} that were not present in smaller models\cite{emergent:LLMs}.
Training these large models requires extensive GPU resources. As a result, a handful of well-funded tech companies and organizations dominate technological advances. This makes LLMs difficult to study from outside. 
Nonetheless, testing and validating technology is critical to ensuring its reliability, safety, and effectiveness, especially for fast-adopted and widely-used technologies like LLMs.

We focus on one of the potential risks of LLMs in generating inaccurate, misleading, false, or entirely fabricated content.  
Even though LLMs are developed with good intentions and applied to perform useful tasks, they can generate factually false statements, known as \textit{hallucinations}\cite{bang2023multitask} and \textit{hallucinatory explanations} in defense of such statements. 
Although widely used, the term ``hallucination'' may be misleading and encourages anthropomorphism\cite{ChatGPTIsntHallucinating}.
Thus, discerning fact from fiction in LLM-generated content is difficult, and it is further complicated by the use of eloquent language and confident tone\cite{doi:10.1080/10447318.2023.2225931}.

The implications of LLM hallucinations and the resulting unsolicited justifications are especially concerning in critical and sensitive fields.  For example, chatbots may become particularly appealing in information-seeking and question-answering tasks related to public health, given that it is easier for many people to receive health advice from a chatbot than from a physician\cite{iftikhar2023docgpt}. Yet, LLMs typically lack access to updated and reliable information sources.
A study involving 6,594 real-world chatbot interactions during the COVID-19 pandemic revealed that 30\% of users queried using the keyword ``COVID-19''\cite{chin2023user}. This is concerning, given the unsuitability of present chatbot systems in addressing emerging topics such as COVID-19, both before the pandemic\cite{palanica_perception_2019} and in hindsight\cite{peskoff-stewart-2023-credible}. A recent expert evaluation revealed that ChatGPT ended up giving a ``definitive answer to something that is not totally agreed on'' on the subject of dialysis for lithium poisoning\cite{peskoff-stewart-2023-credible}. Although advanced agents such as ChatGPT, given their release timing, could potentially provide solid answers to basic health questions with well-established scientific consensus\cite{SRIVASTAVA2021100308}, they are ill-equipped to handle evolving scientific consensus on controversial or emerging topics. 


An even larger concern arises when LLMs are used with malicious intent. There have been several discussions about applications of LLMs that put individuals at risk, such as the generation of inauthentic or misleading content at scale\cite{Menczer2023AI-harms}. 
For example, when ChatGPT was prompted to improve the language of a poorly written phishing email, it not only corrected it but also, without any direct user request, added additional content at the end, asking for money\cite{patel2023creatively}. These capabilities will likely extend beyond mere ``proof-of-concept'' demonstrations, underscoring the risk of propagating false or manipulative information, deliberately or accidentally, with negative consequences for politics, finance, health, and society at large.

LLM-based chatbots also pose challenges to fact-checking. Fact-checkers now need to vet a surge of persuasive yet unreliable text, especially false news generated with the help of LLMs\cite{NewsGuard-Plagiarism-Bot} and spread through social media, including by LLM-based social bots\cite{Yang2023Anatomy-AI-botnet}. There are also questions about whether and how the public can over-rely on chatbots such as ChatGPT as research tools for verifying information\cite{verma2023chatgpt,DeVerna2023AI-factcheck}. Search engines such as Google and Bing have historically been seen as reliable gateways to authoritative information sources; while in certain cases, search engines have helped amplify disinformation\cite{10.1145/3299768}, they are more reliable as they give a clear indication of their sources, unlike ChatGPT.
However, as LLM-based chatbots become increasingly used for information-seeking, there is a risk that the public may receive unreliable information through a modality that has traditionally been trustworthy\cite{verma2023chatgpt}. Recent incidents have highlighted the dangers of this trend, such as Google Bard falsely claiming the James Webb telescope's discovery of an exoplanet image\cite{vincent2023googles} and Replika disseminating misinformation by suggesting that Bill Gates was the architect of COVID-19\cite{marcus2022deep}. These occurrences underscore the delicate balance between the risks and benefits of such advanced technologies.

In summary, LLMs have the potential to mislead the public in novel ways, often in combination with or within environments such as search engines, in which users have come to expect (and assume) accuracy. Moreover, the datasets used to train these models can introduce biases, magnifying specific viewpoints while suppressing others\cite{ferrara2023should}, thus raising concerns about their widespread adoption. This article explores how LLMs like ChatGPT affect fact-checking and truthfulness, emphasizing the tangible risks associated with their hallucinatory capabilities and prospects for malicious use. It also discusses potential opportunities and strategies for addressing these challenges and the limitations of current solutions, concluding with key objectives for individuals, organizations, and governments, advocating the harnessing of LLM benefits while reducing the risks. 


\section{Risks, Threats, and Challenges}


There are two sides to the information veracity problem of LLMs. 
On the one hand, LLMs developed with good intentions may nevertheless be plagued by unreliable information. This could be due to unreliable, inconsistent, incomplete, or missing training data or due to hallucinations.  
As a result, chatbots or AI-enhanced search may give inaccurate responses on sensitive topics such as finance or health, as they may generate false claims or explanations. This is a risk because users may act upon, share, and/or quote these responses later.

On the other hand, AI chatbots may be intentionally used for malicious services or activities, such as scam emails, disinformation websites, or bot feeds. AI-generated fabricated news websites, for example, are known to have tens of thousands of followers on social media, directly reaching social media users worldwide. 
The two sides are not independent of each other; disinformation from low-credibility sources supported by malicious GenAI applications is made available as training data for LLMs\cite{pan2023risk}.
While the focus of this article is on textual output, it is worth noting that AI tools with visual capabilities, such as Dall-E\cite{dalle}, MidJourney\cite{midjourney} and Stable Diffusion\cite{stablediffusion} also pose substantial challenges by enabling the creation of deepfakes and false images at an unprecedented scale\cite{mirsky_deepfake_2021}. Malicious actors can combine fake visuals with AI-generated text to support false claims or to create fake social media profiles on a large scale.

Traditional fact-checking has evolved to ensure the reliability of information\cite{googleCLEF2022CheckThat}, but GenAI poses real threats. The hidden, private nature of chatbot conversations and the potential for misinformation underscore the importance of AI literacy and user awareness. Below, we briefly outline multiple risk factors, imminent threats, and the associated challenges that arise from direct interaction with chatbots and their deliberate malicious use.

\subsection{Factuality Challenges}

The ubiquitous adoption of AI chatbots can aggravate one of our most pressing challenges --- the quality and reliability of information in the digital age. Chatbots can generate false or misleading content that can be quickly disseminated and amplified through various platforms and channels. Some of the main limitations and risks of chatbots are outlined below.

\paragraph{Undersourcing:}
LLM-generated content tends to be coherent but lacks credible sourcing\cite{peskoff-stewart-2023-credible}. This issue extends beyond chatbots to generative search engines. Studies scrutinizing generative search engines also found significant problems with proper citations, with nearly half of the generated sentences lacking citations and only three-quarters of those that have them actually supporting their claims\cite{liu2023evaluating}. These findings raise concerns, considering the growing trust of consumers in their reliability.

\paragraph{Truthfulness:}
Despite their significant strides in natural language generation, LLMs tend to generate undesirable text hallucinations that include nonsensical or significantly divergent output, incoherent content, and factual inaccuracies\cite{fung_hallucinationsurvey_2023}. Although LLMs excel in creative writing and basic explanations, they struggle with factual accuracy. This has led to their temporary ban on platforms such as Stack Overflow\cite{thevergeAIgeneratedAnswers}. Multiple studies highlighted ChatGPT's overall lack of trustworthiness\cite{peskoff-stewart-2023-credible}. In clinical contexts, it demonstrated higher accuracy (80\%) in clinical questions compared to evidence-based ones (36\%)\cite{Kusunose2023}. In another study, ChatGPT was shown to perform better in basic clinical science tests but less effectively in specialized domains, while GPT-4 showed greater consistency between various types of tests\cite{ANTAKI2023100324}. In summary, ChatGPT is not a reliable substitute for actual physicians.
Overall, LLMs are better at deductive reasoning and less so at inductive reasoning\cite{bang2023multitask}, which could affect their reliability for fact-based reasoning tasks\cite{poynterChatGPTFactcheck}.

\paragraph{Confident Tone:}
The text generated by chatbots exudes confidence, which often makes them appear as ``authoritative liars.'' Their persuasive narratives can deceive readers into believing false or meaningless information due to the absence of uncertainty or hedging expressions. Language models tailored for conversation tend to maintain a confident tone even when their accuracy is compromised. Addressing this challenge is difficult because LLMs lack measures of inherent factuality and ways to express related uncertainty. While chatbots such as ChatGPT attempt to emulate uncertainty, most safety checks rely on hard-coded rules based on learned prompts, which can be easily circumvented. 

\paragraph{Fluent Style:}
Chatbots communicate from the first person's perspective and even digress at times like humans being, which encourages anthropomorphism. In addition, the coherent and fluent writing style of chatbots can be persuasive to humans, even on controversial political issues\cite{bai_voelkel_eichstaedt_willer_2023}. 
Numerous studies have shown that fluency shapes the perception of truth ``among intelligent people, despite contradictory knowledge, for claims from unreliable sources'' \cite{brashier2020judging}.
Further studies are needed to examine how users perceive chatbots and their capabilities in terms of AI literacy.

\paragraph{Direct Use:}
In traditional settings, misinformation claims gain prominence and eventually catch the attention of fact-checkers who can debunk them. 
However, chatbot interactions are proprietary and occur privately and without mediation, which makes it harder to discern their validity. 
The misinformation a chatbot conveys may be buried among many other true and false claims, adding to the difficulty of detection and correction. WhatsApp has partnered with a number of fact-checking organisations to create tiplines\cite{WhatsAppTiplines}, where users can manually request verification of suspicious claims. Still, this manual process risks being overwhelmed by GenAI content.

\paragraph{Ease of Access:}
Competition in LLMs is driving fast technological advances. Meta's LLaMA, which came out as a competitor for ChatGPT, is now accessible for personalized use on standard laptop hardware\cite{simonwillisonRunningLLaMA} unlike ChatGPT, which is restricted via a rate-limited API. Recent open-sourcing of models such as LLAMA 2\cite{touvron2023llama}, Falcon\cite{tiiFalcon}, and Jais\cite{jais} offer freely downloadable alternatives 
to encourage transparent and democratized technological development. The ease of access to LLMs can accelerate innovation, but it poses significant risks and challenges when used with malicious intent, as discussed in Section~\ref{threats_posed}. Tracing such malicious use will be more challenging with the growing diversity of available models. 

\paragraph{Halo Effect:}
A model's proficiency in addressing one topic might lead users to believe it can excel in any open-domain conversations, an example of a cognitive bias known as the \emph{halo effect}.\cite{halo}. Such an assumption is risky for users who seek information on new and urgent topics. For example, a chatbot trained before 2019 may not know anything about COVID-19 and can give wrong responses during a health crisis. Even a large, up-to-date training dataset only reflects a tiny and biased selection of human knowledge.

\paragraph{Public Perception:}
An LLM may seem like a reliable ``knowledge base'' to the general public. This can lead some users to consider LLMs as enhanced versions of search engines capable of delivering the best answers. However, this is risky, as demonstrated by a recent incident in which an attorney presented to the court fake cases generated by ChatGPT\cite{forbes:lawyer}. Responses presented in the form of answers, rather than a collection of diverse links offering varying perspectives, may produce either inaccurate or biased information. Therefore, users must learn how LLMs operate and not trust responses blindly. 

\paragraph{Unreliable Evaluation:} 
Another pressing issue is the unreliable evaluation of LLMs\cite{chakraborty2023judging}. Assessing subjective factors such as ``factuality'' and ``truthfulness'' is a complex task beyond predefined and annotated benchmarks such as BIG-bench\cite{srivastava2023beyond}, GLUE\cite{wang-etal-2018-glue}, and SuperGLUE\cite{superglue}. Specialized datasets have been developed to measure the factuality of LLMs, such as TruthfulQA\cite{lin-etal-2022-truthfulqa}, which contains expert-crafted questions that quantify model misconception on the topics of health, law, finance, and politics. There have also been some specialized evaluation measures such as FactScore, which evaluate factuality with respect to Wikipedia (e.g., biography pages)\cite{min2023factscore}.
Benchmark datasets alone are insufficient, as LLMs may have been polluted by encountering the same or similar data during their training \cite{golchin2023time}. Newer metrics such as GPTScore\cite{fu2023gptscore}, G-Eval\cite{liu2023geval}, and SelfCheckGPT\cite{manakul2023selfcheckgpt} attempt to address this shortcoming in evaluations, albeit with limitations due to aspects such as positional and numeric biases, stochastic inferencing, and self-preferencing\cite{wang2023large}. As of now, LLMs have demonstrated mixed performance on misinformation detection tasks\cite{bang2023multitask,DeVerna2023AI-factcheck}, involving test sets consisting of scientific and social claims related to COVID-19\cite{lee-etal-2021-towards}. Maintaining accurate ground-truth references, particularly for factuality assessment, is costly, subject to data drift, and remains an open research problem.

\subsection{Threats Posed by Malicious LLM Usage} \label{threats_posed}

The widespread availability of LLMs empowers people with ideas to craft compelling and elegantly written content and persuasive arguments, a skill traditionally held by experienced writers. Even though all technologies have both good and bad applications, it is important to foresee potential harmful applications of LLMs, as outlined below.

\paragraph{Personalized Attacks:} LLMs can generate text that aligns with the context of the ongoing conversation. This capability can be exploited for malicious purposes, where a user's prior statements, such as those from emails or social media, can be incorporated into a prompt to generate disinformation, phishing messages, harassment or other harmful content on a large scale. The content can be tailored to appear credible, personalized, and targeted at specific individuals or groups and can be mixed with factual statements to make it more persuasive.
The possibility of data leakage increases these risks. 
Cyberhaven, a firm that provides data tracing and security solutions, recently analyzed the engagements of their clients' employees with ChatGPT\cite{cyberhavenDataEmployees} and found that 10.8\% had used ChatGPT at work, 8.6\% had shared company data via prompts, and 4.7\% had accidentally disclosed confidential information. The risk of data leakage is evidenced by the inadvertent disclosure of payment details from 1.2\% of ChatGPT Plus subscribers. 
Open-source LLMs can be manipulated to extract such private information, which can then be used for more effective phishing attacks and other scams.

\paragraph{Style Impersonation:} Fine-tuned LLMs will be able to generate text that emulates the style of any person, providing access to relevant training data. Consequently, it will be relatively simple to train a model to generate text that mimics the writing style of specific individuals, such as journalists, fact-checkers, politicians, regulators, and more. This content could then be distributed on social media platforms in an effort to undermine the credibility of perceived adversaries.

\paragraph{Bypassing Detection:} Fact-checkers prioritize monitoring and verifying widely circulated claims. For example, a piece of misinformation that has circulated a thousand times is scrutinized more than once by only ten people. However, GenAI tools can generate infinite variations of the same content. Even if each variant reaches a small number of people, the cumulative impact could add to that of a highly viral piece while remaining invisible to fact-checkers. This automated diversification could undermine the safety and integrity efforts of many social media platforms, such as Meta's third-party fact-checking program\cite{facebookMetasThirdParty}.

\paragraph{Fake Profiles:} 
The ability to generate credible-sounding fake profiles and content at scale will impact social media influencing, making social bots a more formidable challenge. 
These fake profiles will empower bad actors to infiltrate and manipulate online communities more easily; automation will reduce costs while increasing output quality. 
Fake accounts can be used to spread illicit or manipulative content, such as disinformation and hate speech, creating a skewed perception of popular opinions and norms\cite{truong2023vulnerabilities}. 
We have already observed a large network of ChatGPT-driven fake profiles on Twitter/X\cite{Yang2023Anatomy-AI-botnet}. These deceptive social bots engaged with each other through replies and retweets, created the false appearance of broad support for fake news sources and posted harmful comments while evading detection.  
Social media users are less likely to fact-check and more likely to share false claims when those claims appear to be liked or shared by many others\cite{Fakey2020}. This creates a significant vulnerability to manipulation by fake profiles. Potential harm could even extend to life-threatening matters. For instance, a strong correlation has been reported between the volume of online COVID-19 vaccine misinformation and a reduction in vaccination rates\cite{Pierri2022}. GenAI can be quickly weaponized to amplify such harm at scale.

\section{Addressing the Threats}



Addressing actuality-related challenges in LLMs requires a comprehensive strategy, as no single solution can fully mitigate the adverse consequences. 
Here, we outline various strategic dimensions that, when combined, may lead to more responsible and constructive technological utilization. Some solutions are technological and require building an entirely new LLM. Training a multi-billion-parameter LLM from scratch takes several months and hundreds of GPUs, which is beyond the reach of most academics.  However, smaller models, such as those of LLaMA, Falcon, or Jais, are feasible in academia. For example, running a 7B model needs a single GPU, while a 13B model needs two GPUs.

\paragraph{Alignment and Safety:}
Safety and aligning LLMs with human values and intent have become a major concern for recent models like ChatGPT, LLaMA 2, and Jais. In fact, safety measures are increasingly being considered in all stages of chatbot development --- data cleansing before training the base model, safety instruction-tuning\cite{wang2023donotanswer}, safety in the hidden prompt to the chatbot, and safety in the deployed chatbot via keywords and machine learning.
The availability of open-source LLMs suggests that the effectiveness of alignment efforts, as well as other countermeasures, such as watermarking by large AI companies may be severely limited in mitigating the potential looming threats. Nevertheless, it remains crucial to make every conceivable effort that holds promise in restraining the counterproductive consequences of LLMs. 


\paragraph{Modularized Knowledge Grounded Framework:}
One area where current LLMs significantly fall short is in producing timely, thorough, and well-organized presentations of factually dense information, such as situational and strategic reports. One way to ameliorate this shortcoming in the context of pre-trained LLMs involves a multi-step automated framework for gathering and organizing real-time event information. This modular design can create factually accurate content, which can be further refined using an LLM, as exemplified by SmartBook\cite{reddy2023smartbook}. Initially developed for efficient ground-level reporting during the Russia-Ukraine conflict, SmartBook used LLMs to generate initial situational reports by streamlining event-based timelines, structured summaries, and consolidated references.

\paragraph{Retrieval-Augmented Generation:} 
Retrieval-augmented generation (RAG)\cite{hengji_ketgsurvey_2022,realm2023} incorporates contextual information from external sources into text generation. RAG mitigates the challenge of LLMs producing inaccurate content by enhancing their capabilities with external data. However, it requires efficient retrieval of grounded text at scale and robust evaluation.

\paragraph{Hallucination Control and Knowledge Editing:} 
There are two types of LLM hallucinations~\cite{filippova-2020-controlled}: (i) \textit{faithfulness}, when the generated text is not faithful to the input context; and (ii) \textit{factualness}, when the generated text is not factually correct with respect to world knowledge. Most recent attempts focus on solving the hallucination problem during the inference stage based on an LLM's self-consistency checking~\cite{Gou2023}, cross-model verification~\cite{Cohen2023,Du2023}, or checking against related knowledge~\cite{Dziri2021}. The assumption is that an LLM has knowledge of a given concept, and sampled responses are likely to be similar and contain consistent facts. Another promising line of research focuses on opening up LLMs for knowledge editing. Factual errors can then be localized and fixed by injecting factual updates into the model~\cite{KE2021,MEND2022,SERAC2022,ROME2022,MEMIT2023}. Existing methods focus on factual updates for triples by precise editing. This can be extended to more complex knowledge representations, such as logical rules in the future. 
Another challenge is to evaluate the ``ripple effects'' of knowledge editing in language models. Current knowledge editing benchmarks check that a few paraphrases of the original fact are updated and some unrelated facts are untouched. 
More research must explore whether other facts logically derived from the edit are also changed accordingly.

\paragraph{Alleviating Exposure Bias:}
Exposure bias, i.e., favoring preexisting inductive biases over new ones, persists as a challenge in natural language generation, affecting the output quality of LLMs trained on fixed datasets\cite{schmidt-2019-generalization}. Solutions such as selective upgrade, which dynamically derives relevant instruction-response pairs, aim to address this by improving the ability of LLMs to generalize effectively beyond their training data\cite{yu2023self}.

\paragraph{Better Evaluation:}
Existing evaluation methods such as BERTScore\cite{zhang2020bertscore} and MoverScore\cite{zhao-etal-2019-moverscore} assume similar training and evaluation data distributions, which do not align with the evolving capabilities and requirements of LLMs. This discrepancy is particularly evident in scenarios involving zero-shot instructions and in-context learning, which prominently constitute disparate data distribution scenarios.
Recently proposed evaluation measures such as GPTScore\cite{fu2023gptscore} and G-Eval\cite{liu2023geval} have shown reasonable correlations with human assessments in various tasks, including consistency, accuracy, and correctness. However, weak correlations (around 20--25\%) remain in factuality assessments\cite{fu2023gptscore}, 
suggesting that there is room for improvement. One potential direction is customizing factuality instructions for specific domains, such as medicine or law. Similar adjustments have demonstrated improved factuality assessment in the case of SelfCheckGPT\cite{manakul2023selfcheckgpt}, which is based on the idea that consistently replicable responses are rooted in factual accuracy, as opposed to those generated through stochastic sampling or hallucination, which tend to exhibit more variation.


\paragraph{Privacy and Data Protection:}
While organizations like OpenAI have implemented privacy controls and undergo periodic cybersecurity audits\cite{oaisec}, user studies call for AI systems to adhere to additional data protection regulations, emphasizing steps like data anonymization, aggregation, and differential privacy\cite{datapriv}. While the European Union's AI Act\cite{europaEURLex52021PC0206} has transparency-related obligations based on identified risk levels, stronger regulation might facilitate access to third-party fact-verification APIs/plugins. Chatbot users could then opt-in to use such tools whenever verification is critical to the conversation.

\paragraph{Recognizing AI-Generated Content:}
LLM output is already nearly indistinguishable from human-written text\cite{wang2023m4}. For example, state-of-the-art tools to detect AI-generated content failed to distinguish between legitimate Twitter/X accounts and those managed through ChatGPT\cite{Yang2023Anatomy-AI-botnet}. Furthermore, previous misinformation detectors~\cite{Fung2021} trained from automatically generated fake news training data still perform poorly on detecting human-generated or edited fake content~\cite{misinformation2023}. Future models will likely narrow this gap further, making attempts to label AI-generated text unreliable. Watermarks will also be easy for malicious actors to bypass. Therefore, it is important to study multiple generators in multiple domains, for multiple languages, and using different detectors, and also to maintain constantly growing collections of up-to-date machine-generated content as in the M4 repository\cite{wang2023m4}. 
This issue parallels the rapid advancement of deepfake technology in various media formats, including fake videos, manipulated images, altered audio clips, and stylized text\cite{groh2023human, sadasivan2023aigenerated}. The emergence of sophisticated tactics, such as paraphrasing attacks aimed at evading detection \cite{sadasivan2023aigenerated} and adversarial perturbations resistant to image and video compression codecs\cite{hussain2020adversarial}, underscore the pressing need to address these concerns. 
Devising a robust and fool-proof detection technique, akin to a cat-and-mouse game, remains a formidable challenge.

\paragraph{Content Authenticity and Provenance:}
As AI-generated text becomes ubiquitous, fully human-written text may become more valuable. Technologies and standards for content authenticity and provenance already exist for video and image content\cite{contentauthenticityContentAuthenticity}. These standards describe a way to cryptographically sign content so that the metadata around its creation can be proven not altered. We could use similar methods for text content to prove that they are not AI-generated.
Since AI-generated content can cause harm when it spreads on social media, provenance proofs could be imposed to limit the spread of fake content before it has reached many people\cite{Menczer2023AI-harms}.

\paragraph{Regulation:}
Various regulatory efforts have emerged in light of the disruptive impact of emerging technologies. A new rule in China requires watermarks for AI-generated content, while ChatGPT was temporarily banned in Italy due to GDPR compliance concerns before being finally allowed after due compliance on transparency\cite{chatgptitaly2}. The European Union's AI Act is likely the first to regulate high-risk AI applications spanning multiple sectors\cite{europaEURLex52021PC0206}. In the USA, the Federal Trade Commission has issued warnings against creating misleading tools, emphasizing their prohibition under existing regulations and signaling the potential application of these rules to GenAI\cite{ftcChatbotsDeepfakes}. Along similar lines, the Canadian Directive on Automated Decision-Making has prescribed extensive guidelines that promote data-driven practices and federal compliance, transparency, and reduced negative algorithmic outcomes\cite{canadadir}. One of the challenges to the effectiveness of AI regulation stems from rapid technological advancements. On the one hand, controlling LLMs and their users can be as challenging as handling individuals engaged in phishing and misinformation. On the other hand, bad actors using open-source models will not be bound by regulation.

\paragraph{Public Education:}
Public awareness of deceptively ``slick'' content is essential, similar to our skepticism towards images doctored through Photoshop. This awareness extends to deepfake visual technology. A constructive way for experts to contribute to such education is through resources such as tutorial videos and code,
explaining the basic technology behind ChatGPT. However, raising awareness about LLM-based chatbots is challenging, as the risks they pose demand immediate attention, unlike the gradual understanding of deepfakes for visual content that has developed over the last few years. Another caveat is that skeptical citizens may lose trust in credible, authoritative sources of information, and thus become more vulnerable to conspiracy theories.

\section{Fact-Checking Opportunities}

Leveraging LLMs to foster factuality is as important as addressing the challenges of LLMs.
In fact, one could argue that verifying the truthfulness of a claim may be more important than detecting whether it is AI-generated. 
LLMs present several promising opportunities for fact-checkers and journalists, some of which are highlighted below.
It is important, however, to remain vigilant about the challenges and the ethical considerations associated with such applications of LLMs, including bias, transparency, and the potential for misuse. Balancing the advantages of LLMs with responsible and ethical practices is essential to harness their full potential in domains such as fact-checking.

\paragraph{Fact-Checking Support:}
Although GenAI models lack a concept of truth, they can be used to assist fact-checkers in verifying claims. LLMs can transcribe speeches, debates, interviews, online videos, and news broadcasts, summarize extensive documents, and help create concise lists of crucial claims~\cite{newsclaims2022,sundriyal-etal-2022-empowering,sundriyal2021desyr}. These claims can be further organized into fine-grained claim frames, and the claim-claim relations can be utilized for cross-media cross-lingual fact checking~\cite{Huang2022} and factual error correction~\cite{factualerrorcorrection2023}. This capability is particularly valuable for processing large volumes of news content, enabling fact-checkers to monitor a vast amount of potentially misleading information.
After a fact-checked article has been published, the same claims often reappear. LLMs can help identify sections of documents that repeat a previously fact-checked claim or that make a claim semantically equivalent to a previously verified one. This task does not involve determining the factuality or the falsehood of the claim, as this was already established by fact-checkers or journalists\cite{googleCLEF2022CheckThat}.
Such promising opportunities for LLMs to support fact-checking come with some caveats. First, there are risks of errors when summarizing long documents, such as omitting important context or inserting plausible text that is outside the source material. 
Second, we need to understand how humans will interact with AI-aided fact-checking. A recent study found that fact-checks generated by ChatGPT, even when accurate, did not significantly affect participants' ability to discern headline accuracy or to share accurate news\cite{DeVerna2023AI-factcheck}. Worse, they were harmful in specific cases, such as increasing beliefs in certain false claims. These findings highlight the importance of evaluating unintended consequences of GenAI applications.

\paragraph{Stance Detection:}
There are indications that ChatGPT and other LLMs can help with downstream fact-checking tasks, such as stance detection\cite{zhang2023stance}. On the other hand, whether the LLM has seen the test data, as it was trained on many publicly available datasets remains as a partinent question. Recent studies suggest that ChatGPT is more inclined to excel in stance detection tasks compared to tasks involving emotions or pragmatic analyses, even though it might not have undergone pre-training on evaluation sets explicitly tailored for stance detection tasks\cite{KOCON2023101861}.

\paragraph{Domain-Specific Chat Support:}
There have been attempts to use LLMs to provide a chatbot interface to a domain-specific controlled corpus\cite{itnextRememberingConversations}. Organizations could provide such a service to allow users to search for information on a collection of factually verified articles. Some caution is needed, as LLMs can insert plausible but incorrect information even with a controlled corpus.

\section{Conclusion}


Any tool capable of generating novel content also has the potential to produce misleading content\cite{ferrara2023genai}. Therefore, anyone using LLMs to compose news articles, academic reports, emails, or any text must verify even the most basic facts, regardless of how fluent the text appears. Given the rapid and widespread growth in the use of LLMs, society must act quickly with appropriate regulation, education, and collaboration.
Below, we propose an urgent agenda for individuals, governments, and democratic societies.


\paragraph{Coordination and Collaboration:} 
Towards ensuring the responsible development and deployment of GenAI technologies, nations must make coordinated research investments and establish infrastructure capable of dynamically adapting legislative safeguards, much like the measures in place for other groundbreaking technologies such as human germline editing\cite{nationalacademiesLoginNational}. Moreover, fostering collaborations between political actors and industry leaders is essential to prevent an arms race between AI and AI-detection tools.
By coordinating these efforts globally, we can promote transparent, constructive, and responsible technological development, ensuring that GenAI benefits humanity while minimizing potential risks and abuses.

\paragraph{Regulations:}
Key components of comprehensive regulations in the realm of GenAI include stringent law enforcement to mitigate intentional or inadvertent harm from technology use\cite{bbcChatGPTBanned}, regulatory frameworks for high-risk tech production and sales\cite{ftcChatbotsDeepfakes}, and industry-wide standards for ethical GenAI usage. Special attention should be paid to creating guidelines for journalists who use GenAI and for labeling public ads generated by AI. Adopting standardized evaluation measures tailored to critical sectors will also be vital to ensuring that technology benefits society while minimizing the risks\cite{fu2023gptscore,ANTAKI2023100324}. 

\paragraph{Promoting AI Literacy:}
To promote AI literacy globally, we recommend three key actions: ({i})~conducting AI literacy programs for people of all ages\cite{hussain2020adversarial}, ({ii})~incorporating AI education with an emphasis on ethics into the graduate-level curricula\cite{universityworldnewsUniversityPrinciples}, and ({iii})~sensitizing digital consumers to the potential harms and root causes of GenAI\cite{genAIeducationUK}. These measures will empower individuals to engage with AI responsibly and safely, fostering a more knowledgeable and aware global community.

\paragraph{Technological Development:} As common R$\&$D best practices, we recommend integrating clear and accessible informative material for users upfront to clarify the limits (e.g., in terms of factuality) and the risks (e.g., because of the
``authoritative'' tone) of LLMs. To enhance safety, robust guardrails should be implemented in LLM-based conversations\cite{nvidiaRightTrack}. Chatbots should be equipped with evidence-supporting capabilities\cite{Chen2023-fj} to bolster credibility, while external knowledge sources like retrieval-augmented LLMs and knowledge graphs can ensure factual consistency\cite{microsoftIntroducingMicrosoft}. Lastly, the development of coordination detection algorithms is essential to identify suspiciously similar and potentially harmful content\cite{Pacheco2021Coordinated} generated by malicious actors using LLMs\cite{Menczer2023AI-harms}.

\section*{Author Contributions} 
All the authors contributed to conceptualizing, preparing, and finalizing the manuscript. The author names are arranged alphabetically by the last names.

\section*{Funding Information}
M.C. is supported by the Institute for Basic Science (grant IBS-R029-C2) and the National Research Foundation of Korea (grant RS-2022-00165347). 
T.C. acknowledges the financial support of Wipro AI. 
I.A. is supported in part by the European Union (ERC, ExplainYourself, 101077481). 
G.L.C. is supported by the National Science Foundation (grant 2239194). 
E.F. and F.M. are partly supported by DARPA (grant HR001121C0169). 
F.M. is also partly supported by the Knight Foundation and Craig Newmark Philanthropies. 
G.Z.’s fact-checking project receives funding from the European Union through multiple grants and is part of Meta's 3PFC Program. 
H.J.'s research is partially supported by U.S. DARPA SemaFor Program No. HR001120C0123. 
Any opinions, findings, conclusions, or recommendations expressed in this material are those of the authors and do not necessarily reflect the views of the funders.

\section*{Competing Interests}
The authors declare no competing interests.

\section*{Additional Information}

\noindent{\bf Materials \& Correspondence} should be emailed to Tanmoy Chakraborty (\url{tanchak@iitd.ac.in}).

\bibliography{sample,sn-bibliography,giovanni,keyrefs_arxiv}

\begin{thebibliography}{100}
\urlstyle{rm}
\expandafter\ifx\csname url\endcsname\relax
  \def\url#1{\texttt{#1}}\fi
\expandafter\ifx\csname urlprefix\endcsname\relax\def\urlprefix{URL }\fi
\expandafter\ifx\csname doiprefix\endcsname\relax\def\doiprefix{DOI: }\fi
\providecommand{\bibinfo}[2]{#2}
\providecommand{\eprint}[2][]{\url{#2}}

\bibitem{shannon1948mathematical}
\bibinfo{author}{Shannon, C.~E.}
\newblock \bibinfo{journal}{\bibinfo{title}{A mathematical theory of
  communication}}.
\newblock {\emph{\JournalTitle{The Bell System Technical Journal}}}
  \textbf{\bibinfo{volume}{27}}, \bibinfo{pages}{379--423},
  \doiprefix\url{10.1002/j.1538-7305.1948.tb01338.x} (\bibinfo{year}{1948}).

\bibitem{huang-chang-2023-towards}
\bibinfo{author}{Huang, J.} \& \bibinfo{author}{Chang, K. C.-C.}
\newblock \bibinfo{title}{{Towards Reasoning in Large Language Models: A
  Survey}}.
\newblock In \emph{\bibinfo{booktitle}{Findings of the Association for
  Computational Linguistics: ACL 2023}}, \bibinfo{pages}{1049--1065},
  \doiprefix\url{10.18653/v1/2023.findings-acl.67}
  (\bibinfo{publisher}{Association for Computational Linguistics},
  \bibinfo{address}{Toronto, Canada}, \bibinfo{year}{2023}).

\bibitem{radford2018improving}
\bibinfo{author}{Radford, A.}, \bibinfo{author}{Narasimhan, K.},
  \bibinfo{author}{Salimans, T.}, \bibinfo{author}{Sutskever, I.} \emph{et~al.}
\newblock \bibinfo{title}{Improving language understanding by generative
  pre-training} (\bibinfo{year}{2018}).

\bibitem{openai2023gpt4}
\bibinfo{author}{OpenAI}.
\newblock \bibinfo{journal}{\bibinfo{title}{{GPT}-4 technical report}}.
\newblock {\emph{\JournalTitle{arXiv:2303.08774}}}  (\bibinfo{year}{2023}).

\bibitem{touvron2023llama}
\bibinfo{author}{Touvron, H.} \emph{et~al.}
\newblock \bibinfo{journal}{\bibinfo{title}{Llama: Open and efficient
  foundation language models}}.
\newblock {\emph{\JournalTitle{arXiv preprint arXiv:2302.13971}}}
  (\bibinfo{year}{2023}).

\bibitem{openaiIntroducingChatGPT}
\bibinfo{title}{{I}ntroducing {C}hat{G}{P}{T} --- openai.com}.
\newblock \bibinfo{howpublished}{\url{https://openai.com/blog/chatgpt}}.
\newblock \bibinfo{note}{[Accessed 27-09-2023]}.

\bibitem{hu2023chatgpt}
\bibinfo{author}{Hu, K.}
\newblock \bibinfo{journal}{\bibinfo{title}{{ChatGPT} sets record for
  fastest-growing user base - analyst note}}.
\newblock {\emph{\JournalTitle{Reuters}}}  (\bibinfo{year}{2023}).

\bibitem{thoppilan2022lamda}
\bibinfo{author}{Thoppilan, R.} \emph{et~al.}
\newblock \bibinfo{journal}{\bibinfo{title}{Lamda: Language models for dialog
  applications}}.
\newblock {\emph{\JournalTitle{arXiv preprint arXiv:2201.08239}}}
  (\bibinfo{year}{2022}).

\bibitem{alpaca}
\bibinfo{author}{Taori, R.} \emph{et~al.}
\newblock \bibinfo{title}{Stanford {A}lpaca: An instruction-following {LLaMA}
  model}.
\newblock
  \bibinfo{howpublished}{\url{https://github.com/tatsu-lab/stanford_alpaca}}
  (\bibinfo{year}{2023}).

\bibitem{vicuna}
\bibinfo{title}{Vicuna: An open-source chatbot impressing {GPT}-4 with 90\%*
  {ChatGPT} quality}.
\newblock
  \bibinfo{howpublished}{\url{https://lmsys.org/blog/2023-03-30-vicuna/}}.
\newblock \bibinfo{note}{[Accessed 15-09-2023]}.

\bibitem{claude}
\bibinfo{title}{Introducing {C}laude}.
\newblock
  \bibinfo{howpublished}{\url{https://www.anthropic.com/index/introducing-claude}}.
\newblock \bibinfo{note}{[Accessed 15-09-2023]}.

\bibitem{falcon}
\bibinfo{title}{Falcon}.
\newblock \bibinfo{howpublished}{\url{https://falconllm.tii.ae/falcon.html}}.
\newblock \bibinfo{note}{[Accessed 15-09-2023]}.

\bibitem{jurassic}
\bibinfo{title}{Jurassic-2 models}.
\newblock
  \bibinfo{howpublished}{\url{https://docs.ai21.com/docs/jurassic-2-models}}.
\newblock \bibinfo{note}{[Accessed 20-09-2023]}.

\bibitem{jais}
\bibinfo{author}{Sengupta, N.} \emph{et~al.}
\newblock \bibinfo{journal}{\bibinfo{title}{Jais and {J}ais-chat:
  {A}rabic-centric foundation and instruction-tuned open generative large
  language models}}.
\newblock {\emph{\JournalTitle{arXiv preprint 2308.16149}}}
  (\bibinfo{year}{2023}).
\newblock \eprint{2308.16149}.

\bibitem{radford2019language}
\bibinfo{author}{Radford, A.} \emph{et~al.}
\newblock \bibinfo{journal}{\bibinfo{title}{Language models are unsupervised
  multitask learners}}.
\newblock {\emph{\JournalTitle{OpenAI blog}}} \textbf{\bibinfo{volume}{1}},
  \bibinfo{pages}{9} (\bibinfo{year}{2019}).

\bibitem{brown2020language}
\bibinfo{author}{Brown, T.} \emph{et~al.}
\newblock \bibinfo{title}{Language {Models} are {Few}-{Shot} {Learners}}.
\newblock In \emph{\bibinfo{booktitle}{Advances in {Neural} {Information}
  {Processing} {Systems}}}, vol.~\bibinfo{volume}{33},
  \bibinfo{pages}{1877--1901} (\bibinfo{publisher}{Curran Associates, Inc.},
  \bibinfo{year}{2020}).

\bibitem{LLMSurvey}
\bibinfo{author}{Zhao, W.~X.} \emph{et~al.}
\newblock \bibinfo{journal}{\bibinfo{title}{A survey of large language
  models}}.
\newblock {\emph{\JournalTitle{arXiv preprint arXiv:2303.18223}}}
  (\bibinfo{year}{2023}).

\bibitem{hierarchicalschema2023}
\bibinfo{author}{Li, S.} \emph{et~al.}
\newblock \bibinfo{title}{Open-domain hierarchical event schema induction by
  incremental prompting and verification}.
\newblock In \emph{\bibinfo{booktitle}{Proceedings of the 61st Annual Meeting
  of the Association for Computational Linguistics (Volume 1: Long Papers)}},
  \bibinfo{pages}{5677--5697}, \doiprefix\url{10.18653/v1/2023.acl-long.312}
  (\bibinfo{publisher}{Association for Computational Linguistics},
  \bibinfo{address}{Toronto, Canada}, \bibinfo{year}{2023}).

\bibitem{emergent:LLMs}
\bibinfo{author}{Wei, J.} \emph{et~al.}
\newblock \bibinfo{journal}{\bibinfo{title}{Emergent abilities of large
  language models}}.
\newblock {\emph{\JournalTitle{Trans. Mach. Learn. Res.}}}
  \textbf{\bibinfo{volume}{2022}} (\bibinfo{year}{2022}).

\bibitem{bang2023multitask}
\bibinfo{author}{Bang, Y.} \emph{et~al.}
\newblock \bibinfo{title}{{A Multitask, Multilingual, Multimodal Evaluation of
  ChatGPT on Reasoning, Hallucination, and Interactivity}}.
\newblock \bibinfo{type}{Preprint} \bibinfo{number}{2302.04023},
  \bibinfo{institution}{arXiv} (\bibinfo{year}{2023}).
\newblock \doiprefix\url{10.48550/arXiv.2302.04023}.

\bibitem{ChatGPTIsntHallucinating}
\bibinfo{author}{Bergstrom, C.~T.} \& \bibinfo{author}{Ogbunu, C.~B.}
\newblock \bibinfo{title}{{ChatGPT} isn’t ‘hallucinating.’ {It’s
  Bullshitting}.}
\newblock
  \bibinfo{howpublished}{\url{https://undark.org/2023/04/06/chatgpt-isnt-hallucinating-its-bullshitting/}}
  (\bibinfo{year}{2023}).
\newblock \bibinfo{note}{Accessed 2 October 2023}.

\bibitem{doi:10.1080/10447318.2023.2225931}
\bibinfo{author}{Sison, A. J.~G.}, \bibinfo{author}{Daza, M.~T.},
  \bibinfo{author}{Gozalo-Brizuela, R.} \&
  \bibinfo{author}{Garrido-Merch{\'a}n, E.~C.}
\newblock \bibinfo{journal}{\bibinfo{title}{{ChatGPT: More Than a ``Weapon of
  Mass Deception'' -- Ethical Challenges and Responses from the Human-Centered
  Artificial Intelligence (HCAI) Perspective}}}.
\newblock {\emph{\JournalTitle{International Journal of Human-Computer
  Interaction}}} \textbf{\bibinfo{volume}{0}}, \bibinfo{pages}{1--20},
  \doiprefix\url{10.1080/10447318.2023.2225931} (\bibinfo{year}{2023}).
\newblock \eprint{https://doi.org/10.1080/10447318.2023.2225931}.

\bibitem{iftikhar2023docgpt}
\bibinfo{author}{Iftikhar, L.} \emph{et~al.}
\newblock \bibinfo{journal}{\bibinfo{title}{Docgpt: Impact of chatgpt-3 on
  health services as a virtual doctor}}.
\newblock {\emph{\JournalTitle{EC Paediatrics}}} \textbf{\bibinfo{volume}{12}},
  \bibinfo{pages}{45--55} (\bibinfo{year}{2023}).

\bibitem{chin2023user}
\bibinfo{author}{Chin, H.} \emph{et~al.}
\newblock \bibinfo{journal}{\bibinfo{title}{User-chatbot conversations during
  the {COVID}-19 pandemic: Study based on topic modeling and sentiment
  analysis}}.
\newblock {\emph{\JournalTitle{Journal of Medical Internet Research}}}
  \textbf{\bibinfo{volume}{25}}, \bibinfo{pages}{e40922},
  \doiprefix\url{10.2196/40922} (\bibinfo{year}{2023}).

\bibitem{palanica_perception_2019}
\bibinfo{author}{Palanica, A.}, \bibinfo{author}{Flaschner, P.},
  \bibinfo{author}{Thommandram, A.}, \bibinfo{author}{Li, M.} \&
  \bibinfo{author}{Fossat, Y.}
\newblock \bibinfo{journal}{\bibinfo{title}{Physicians' perceptions of chatbots
  in health care: Cross-sectional web-based survey}}.
\newblock {\emph{\JournalTitle{J Med Internet Res}}}
  \textbf{\bibinfo{volume}{21}}, \bibinfo{pages}{e12887},
  \doiprefix\url{10.2196/12887} (\bibinfo{year}{2019}).

\bibitem{peskoff-stewart-2023-credible}
\bibinfo{author}{Peskoff, D.} \& \bibinfo{author}{Stewart, B.}
\newblock \bibinfo{title}{Credible without credit: Domain experts assess
  generative language models}.
\newblock In \emph{\bibinfo{booktitle}{Proceedings of the 61st Annual Meeting
  of the Association for Computational Linguistics (Volume 2: Short Papers)}},
  \bibinfo{pages}{427--438}, \doiprefix\url{10.18653/v1/2023.acl-short.37}
  (\bibinfo{publisher}{Association for Computational Linguistics},
  \bibinfo{address}{Toronto, Canada}, \bibinfo{year}{2023}).

\bibitem{SRIVASTAVA2021100308}
\bibinfo{author}{Srivastava, B.}
\newblock \bibinfo{journal}{\bibinfo{title}{Did chatbots miss their “{A}pollo
  moment”? {P}otential, gaps, and lessons from using collaboration assistants
  during {COVID}-19}}.
\newblock {\emph{\JournalTitle{Patterns}}} \textbf{\bibinfo{volume}{2}},
  \bibinfo{pages}{100308},
  \doiprefix\url{https://doi.org/10.1016/j.patter.2021.100308}
  (\bibinfo{year}{2021}).

\bibitem{Menczer2023AI-harms}
\bibinfo{author}{Menczer, F.}, \bibinfo{author}{Crandall, D.},
  \bibinfo{author}{Ahn, Y.-Y.} \& \bibinfo{author}{Kapadia, A.}
\newblock \bibinfo{journal}{\bibinfo{title}{{Addressing the harms of
  AI-generated inauthentic content}}}.
\newblock {\emph{\JournalTitle{Nature Machine Intelligence}}}
  \textbf{\bibinfo{volume}{5}}, \bibinfo{pages}{678--680},
  \doiprefix\url{10.1038/s42256-023-00690-w} (\bibinfo{year}{2023}).

\bibitem{patel2023creatively}
\bibinfo{author}{Patel, A.} \& \bibinfo{author}{Sattler, J.}
\newblock \bibinfo{title}{Creatively malicious prompt engineering}.
\newblock \bibinfo{type}{Tech. Rep.}, \bibinfo{institution}{{WithSecure} Labs}
  (\bibinfo{year}{2023}).

\bibitem{NewsGuard-Plagiarism-Bot}
\bibinfo{author}{Brewster, J.}, \bibinfo{author}{Wang, M.} \&
  \bibinfo{author}{Palmer, C.}
\newblock \bibinfo{title}{{Plagiarism-Bot? How Low-Quality Websites Are Using
  AI to Deceptively Rewrite Content from Mainstream News Outlets}}.
\newblock
  \bibinfo{howpublished}{\url{https://www.newsguardtech.com/misinformation-monitor/august-2023/}}
  (\bibinfo{year}{2023}).
\newblock \bibinfo{note}{[Accessed 23 Sep 2023]}.

\bibitem{Yang2023Anatomy-AI-botnet}
\bibinfo{author}{Yang, K.-C.} \& \bibinfo{author}{Menczer, F.}
\newblock \bibinfo{title}{{Anatomy of an AI-powered malicious social botnet}}.
\newblock \bibinfo{type}{Preprint} \bibinfo{number}{2307.16336},
  \bibinfo{institution}{arXiv} (\bibinfo{year}{2023}).
\newblock \doiprefix\url{10.48550/arXiv.2307.16336}.

\bibitem{verma2023chatgpt}
\bibinfo{author}{Verma, P.} \& \bibinfo{author}{Oremus, W.}
\newblock \bibinfo{journal}{\bibinfo{title}{{ChatGPT} invented a sexual
  harassment scandal and named a real law prof as the accused}}.
\newblock {\emph{\JournalTitle{Washington Post}}}  (\bibinfo{year}{2023}).

\bibitem{DeVerna2023AI-factcheck}
\bibinfo{author}{DeVerna, M.~R.}, \bibinfo{author}{Yan, H.~Y.},
  \bibinfo{author}{Yang, K.-C.} \& \bibinfo{author}{Menczer, F.}
\newblock \bibinfo{title}{Artificial intelligence is ineffective and
  potentially harmful for fact checking}.
\newblock \bibinfo{type}{Preprint} \bibinfo{number}{2308.10800},
  \bibinfo{institution}{arXiv} (\bibinfo{year}{2023}).
\newblock \doiprefix\url{10.48550/arXiv.2308.10800}.

\bibitem{10.1145/3299768}
\bibinfo{author}{Ferrara, E.}
\newblock \bibinfo{journal}{\bibinfo{title}{The history of digital spam}}.
\newblock {\emph{\JournalTitle{Commun. ACM}}} \textbf{\bibinfo{volume}{62}},
  \bibinfo{pages}{82–91}, \doiprefix\url{10.1145/3299768}
  (\bibinfo{year}{2019}).

\bibitem{vincent2023googles}
\bibinfo{author}{Vincent, J.}
\newblock \bibinfo{title}{Google’s {AI} chatbot {Bard} makes factual error in
  first demo} (\bibinfo{year}{2023}).

\bibitem{marcus2022deep}
\bibinfo{author}{Marcus, G.}
\newblock \bibinfo{title}{Deep {Learning} {Is} {Hitting} a {Wall}}
  (\bibinfo{year}{2022}).

\bibitem{ferrara2023should}
\bibinfo{author}{Ferrara, E.}
\newblock \bibinfo{journal}{\bibinfo{title}{{Should ChatGPT be Biased?
  Challenges and Risks of Bias in Large Language Models}}}.
\newblock {\emph{\JournalTitle{Should ChatGPT be Biased? Challenges and Risks
  of Bias in Large Language Models}}}  (\bibinfo{year}{2023}).

\bibitem{pan2023risk}
\bibinfo{author}{Pan, Y.} \emph{et~al.}
\newblock \bibinfo{journal}{\bibinfo{title}{On the risk of misinformation
  pollution with large language models}}.
\newblock {\emph{\JournalTitle{arXiv preprint 2305.13661}}}
  (\bibinfo{year}{2023}).
\newblock \eprint{2305.13661}.

\bibitem{dalle}
\bibinfo{title}{Dall.e 2}.
\newblock \bibinfo{howpublished}{\url{https://openai.com/dall-e-2}}.
\newblock \bibinfo{note}{[Accessed 15-09-2023]}.

\bibitem{midjourney}
\bibinfo{title}{Midjourney}.
\newblock \bibinfo{howpublished}{\url{https://www.midjourney.com/}}.
\newblock \bibinfo{note}{[Accessed 15-09-2023]}.

\bibitem{stablediffusion}
\bibinfo{title}{Stable diffusion}.
\newblock \bibinfo{howpublished}{\url{https://stablediffusionweb.com/}}.
\newblock \bibinfo{note}{[Accessed 15-09-2023]}.

\bibitem{mirsky_deepfake_2021}
\bibinfo{author}{Mirsky, Y.} \& \bibinfo{author}{Lee, W.}
\newblock \bibinfo{journal}{\bibinfo{title}{{The Creation and Detection of
  Deepfakes: A Survey}}}.
\newblock {\emph{\JournalTitle{ACM Comput. Surv.}}}
  \textbf{\bibinfo{volume}{54}}, \doiprefix\url{10.1145/3425780}
  (\bibinfo{year}{2021}).

\bibitem{googleCLEF2022CheckThat}
\bibinfo{title}{{C}{L}{E}{F}2022-{C}heck{T}hat! --- sites.google.com}.
\newblock
  \bibinfo{howpublished}{\url{https://sites.google.com/view/clef2022-checkthat}}.

\bibitem{liu2023evaluating}
\bibinfo{author}{Liu, N.~F.}, \bibinfo{author}{Zhang, T.} \&
  \bibinfo{author}{Liang, P.}
\newblock \bibinfo{title}{Evaluating verifiability in generative search
  engines} (\bibinfo{year}{2023}).
\newblock \eprint{2304.09848}.

\bibitem{fung_hallucinationsurvey_2023}
\bibinfo{author}{Ji, Z.} \emph{et~al.}
\newblock \bibinfo{journal}{\bibinfo{title}{Survey of hallucination in natural
  language generation}}.
\newblock {\emph{\JournalTitle{ACM Comput. Surv.}}}
  \textbf{\bibinfo{volume}{55}}, \doiprefix\url{10.1145/3571730}
  (\bibinfo{year}{2023}).

\bibitem{thevergeAIgeneratedAnswers}
\bibinfo{author}{Vincent, J.}
\newblock \bibinfo{title}{{AI}-generated answers temporarily banned on coding
  {Q}\&{A} site {S}tack {O}verflow --- theverge.com}.
\newblock
  \bibinfo{howpublished}{\url{https://www.theverge.com/2022/12/5/23493932/chatgpt-ai-generated-answers-temporarily-banned-stack-overflow-llms-dangers}}.
\newblock \bibinfo{note}{[Accessed 08-09-2023]}.

\bibitem{Kusunose2023}
\bibinfo{author}{Kusunose, K.}, \bibinfo{author}{Kashima, S.} \&
  \bibinfo{author}{Sata, M.}
\newblock \bibinfo{journal}{\bibinfo{title}{Evaluation of the accuracy of
  {ChatGPT} in answering clinical questions on the {Japanese Society of
  Hypertension Guidelines}}}.
\newblock {\emph{\JournalTitle{Circulation Journal}}}
  \textbf{\bibinfo{volume}{87}}, \bibinfo{pages}{1030--1033},
  \doiprefix\url{10.1253/circj.CJ-23-0308} (\bibinfo{year}{2023}).

\bibitem{ANTAKI2023100324}
\bibinfo{author}{Antaki, F.}, \bibinfo{author}{Touma, S.},
  \bibinfo{author}{Milad, D.}, \bibinfo{author}{El-Khoury, J.} \&
  \bibinfo{author}{Duval, R.}
\newblock \bibinfo{journal}{\bibinfo{title}{{Evaluating the Performance of
  {ChatGPT} in Ophthalmology: An Analysis of Its Successes and Shortcomings}}}.
\newblock {\emph{\JournalTitle{Ophthalmology Science}}}
  \textbf{\bibinfo{volume}{3}}, \bibinfo{pages}{100324},
  \doiprefix\url{https://doi.org/10.1016/j.xops.2023.100324}
  (\bibinfo{year}{2023}).

\bibitem{poynterChatGPTFactcheck}
\bibinfo{author}{Abels, G.}
\newblock \bibinfo{title}{{C}an {C}hat{G}{P}{T} fact-check? {W}e tested. -
  {P}oynter --- poynter.org}.
\newblock
  \bibinfo{howpublished}{\url{https://www.poynter.org/fact-checking/2023/chatgpt-ai-replace-fact-checking/}}.
\newblock \bibinfo{note}{[Accessed 27-09-2023]}.

\bibitem{bai_voelkel_eichstaedt_willer_2023}
\bibinfo{author}{Bai, H.}, \bibinfo{author}{Voelkel, J.~G.},
  \bibinfo{author}{Eichstaedt, j.~C.} \& \bibinfo{author}{Willer, R.}
\newblock \bibinfo{title}{Artificial intelligence can persuade humans on
  political issues}, \doiprefix\url{10.31219/osf.io/stakv}
  (\bibinfo{year}{2023}).

\bibitem{brashier2020judging}
\bibinfo{author}{Brashier, N.~M.} \& \bibinfo{author}{Marsh, E.~J.}
\newblock \bibinfo{journal}{\bibinfo{title}{Judging truth}}.
\newblock {\emph{\JournalTitle{Annual review of psychology}}}
  \textbf{\bibinfo{volume}{71}}, \bibinfo{pages}{499--515}
  (\bibinfo{year}{2020}).

\bibitem{WhatsAppTiplines}
\bibinfo{title}{{IFCN} fact-checking organizations on {WhatsApp}}.
\newblock
  \bibinfo{howpublished}{\url{https://faq.whatsapp.com/5059120540855664}}
  (\bibinfo{year}{2023}).
\newblock \bibinfo{note}{Accessed 2 October 2023}.

\bibitem{simonwillisonRunningLLaMA}
\bibinfo{title}{{R}unning {L}{L}a{M}{A} 7{B} and 13{B} on a 64{G}{B} {M}2
  {M}ac{B}ook {P}ro with llama.cpp --- til.simonwillison.net}.
\newblock
  \bibinfo{howpublished}{\url{https://til.simonwillison.net/llms/llama-7b-m2}}.
\newblock \bibinfo{note}{[Accessed 08-09-2023]}.

\bibitem{tiiFalcon}
\bibinfo{title}{{F}alcon {L}{L}{M} --- falconllm.tii.ae}.
\newblock
  \bibinfo{howpublished}{\url{https://falconllm.tii.ae/falcon-180b.html}}.
\newblock \bibinfo{note}{[Accessed 08-09-2023]}.

\bibitem{halo}
\bibinfo{author}{Volunteering, C.}
\newblock \bibinfo{title}{Halo effect --- wikipedia.org}.
\newblock
  \bibinfo{howpublished}{\url{https://en.wikipedia.org/wiki/Halo_effect}}.
\newblock \bibinfo{note}{[Accessed 27-09-2023]}.

\bibitem{forbes:lawyer}
\bibinfo{author}{OpenAI}.
\newblock \bibinfo{title}{Lawyer used {ChatGPT} in court—and cited fake
  cases. {A} judge is considering sanctions}.
\newblock
  \bibinfo{howpublished}{\url{https://www.forbes.com/sites/mollybohannon/2023/06/08/lawyer-used-chatgpt-in-court-and-cited-fake-cases-a-judge-is-considering-sanctions/}}.
\newblock \bibinfo{note}{[Accessed 15-09-2023]}.

\bibitem{chakraborty2023judging}
\bibinfo{author}{Chakraborty, T.} \& \bibinfo{author}{Masud, S.}
\newblock \bibinfo{journal}{\bibinfo{title}{Judging the creative prowess of
  {AI}}}.
\newblock {\emph{\JournalTitle{Nature Machine Intelligence}}}
  \bibinfo{pages}{1--1} (\bibinfo{year}{2023}).

\bibitem{srivastava2023beyond}
\bibinfo{author}{Srivastava, A.} \emph{et~al.}
\newblock \bibinfo{journal}{\bibinfo{title}{{Beyond the Imitation Game:
  Quantifying and extrapolating the capabilities of language models}}}.
\newblock {\emph{\JournalTitle{Transactions on Machine Learning Research}}}
  (\bibinfo{year}{2023}).

\bibitem{wang-etal-2018-glue}
\bibinfo{author}{Wang, A.} \emph{et~al.}
\newblock \bibinfo{title}{{GLUE}: A multi-task benchmark and analysis platform
  for natural language understanding}.
\newblock In \emph{\bibinfo{booktitle}{Proceedings of the 2018 {EMNLP} Workshop
  {B}lackbox{NLP}: Analyzing and Interpreting Neural Networks for {NLP}}},
  \bibinfo{pages}{353--355}, \doiprefix\url{10.18653/v1/W18-5446}
  (\bibinfo{publisher}{Association for Computational Linguistics},
  \bibinfo{address}{Brussels, Belgium}, \bibinfo{year}{2018}).

\bibitem{superglue}
\bibinfo{author}{Wang, A.} \emph{et~al.}
\newblock \bibinfo{title}{{SuperGLUE: A Stickier Benchmark for General-Purpose
  Language Understanding Systems}}.
\newblock In \emph{\bibinfo{booktitle}{Proceedings of the 33rd International
  Conference on Neural Information Processing Systems}}, \bibinfo{pages}{294}
  (\bibinfo{publisher}{Curran Associates Inc.}, \bibinfo{address}{Red Hook, NY,
  USA}, \bibinfo{year}{2019}).

\bibitem{lin-etal-2022-truthfulqa}
\bibinfo{author}{Lin, S.}, \bibinfo{author}{Hilton, J.} \&
  \bibinfo{author}{Evans, O.}
\newblock \bibinfo{title}{{T}ruthful{QA}: Measuring how models mimic human
  falsehoods}.
\newblock In \emph{\bibinfo{booktitle}{Proceedings of the 60th Annual Meeting
  of the Association for Computational Linguistics (Volume 1: Long Papers)}},
  \bibinfo{pages}{3214--3252}, \doiprefix\url{10.18653/v1/2022.acl-long.229}
  (\bibinfo{publisher}{Association for Computational Linguistics},
  \bibinfo{address}{Dublin, Ireland}, \bibinfo{year}{2022}).

\bibitem{min2023factscore}
\bibinfo{author}{Min, S.} \emph{et~al.}
\newblock \bibinfo{journal}{\bibinfo{title}{{FActScore: Fine-grained Atomic
  Evaluation of Factual Precision in Long Form Text Generation}}}.
\newblock {\emph{\JournalTitle{arXiv preprint 2305.14251}}}
  (\bibinfo{year}{2023}).
\newblock \eprint{2305.14251}.

\bibitem{golchin2023time}
\bibinfo{author}{Golchin, S.} \& \bibinfo{author}{Surdeanu, M.}
\newblock \bibinfo{title}{{Time Travel in LLMs: Tracing Data Contamination in
  Large Language Models}} (\bibinfo{year}{2023}).
\newblock \eprint{2308.08493}.

\bibitem{fu2023gptscore}
\bibinfo{author}{Fu, J.}, \bibinfo{author}{Ng, S.-K.}, \bibinfo{author}{Jiang,
  Z.} \& \bibinfo{author}{Liu, P.}
\newblock \bibinfo{title}{{GPTScore}: Evaluate as you desire}
  (\bibinfo{year}{2023}).
\newblock \eprint{2302.04166}.

\bibitem{liu2023geval}
\bibinfo{author}{Liu, Y.} \emph{et~al.}
\newblock \bibinfo{title}{{G-Eval}: {NLG} evaluation using {GPT}-4 with better
  human alignment} (\bibinfo{year}{2023}).
\newblock \eprint{2303.16634}.

\bibitem{manakul2023selfcheckgpt}
\bibinfo{author}{Manakul, P.}, \bibinfo{author}{Liusie, A.} \&
  \bibinfo{author}{Gales, M. J.~F.}
\newblock \bibinfo{title}{{SelfCheckGPT}: Zero-resource black-box hallucination
  detection for generative large language models} (\bibinfo{year}{2023}).
\newblock \eprint{2303.08896}.

\bibitem{wang2023large}
\bibinfo{author}{Wang, P.} \emph{et~al.}
\newblock \bibinfo{title}{Large language models are not fair evaluators}
  (\bibinfo{year}{2023}).
\newblock \eprint{2305.17926}.

\bibitem{lee-etal-2021-towards}
\bibinfo{author}{Lee, N.}, \bibinfo{author}{Bang, Y.},
  \bibinfo{author}{Madotto, A.} \& \bibinfo{author}{Fung, P.}
\newblock \bibinfo{title}{Towards few-shot fact-checking via perplexity}.
\newblock In \emph{\bibinfo{booktitle}{Proceedings of the 2021 Conference of
  the North American Chapter of the Association for Computational Linguistics:
  Human Language Technologies}}, \bibinfo{pages}{1971--1981},
  \doiprefix\url{10.18653/v1/2021.naacl-main.158}
  (\bibinfo{publisher}{Association for Computational Linguistics},
  \bibinfo{address}{Online}, \bibinfo{year}{2021}).

\bibitem{cyberhavenDataEmployees}
\bibinfo{author}{Coles, C.}
\newblock \bibinfo{title}{11\% of data employees paste into {C}hat{G}{P}{T} is
  confidential - {C}yberhaven --- cyberhaven.com}.
\newblock
  \bibinfo{howpublished}{\url{https://www.cyberhaven.com/blog/4-2-of-workers-have-pasted-company-data-into-chatgpt/}}.
\newblock \bibinfo{note}{[Accessed 08-09-2023]}.

\bibitem{facebookMetasThirdParty}
\bibinfo{title}{{M}eta's {T}hird-{P}arty {F}act-{C}hecking {P}rogram | {M}eta
  {J}ournalism {P}roject --- facebook.com}.
\newblock
  \bibinfo{howpublished}{\url{https://www.facebook.com/formedia/mjp/programs/third-party-fact-checking}}.

\bibitem{truong2023vulnerabilities}
\bibinfo{author}{Truong, B.~T.}, \bibinfo{author}{Lou, X.},
  \bibinfo{author}{Flammini, A.} \& \bibinfo{author}{Menczer, F.}
\newblock \bibinfo{title}{Vulnerabilities of the online public square to
  manipulation}.
\newblock \bibinfo{type}{Preprint} \bibinfo{number}{1907.06130},
  \bibinfo{institution}{arXiv} (\bibinfo{year}{2023}).
\newblock \doiprefix\url{10.48550/arXiv.1907.06130}.

\bibitem{Fakey2020}
\bibinfo{author}{Avram, M.}, \bibinfo{author}{Micallef, N.},
  \bibinfo{author}{Patil, S.} \& \bibinfo{author}{Menczer, F.}
\newblock \bibinfo{journal}{\bibinfo{title}{Exposure to social engagement
  metrics increases vulnerability to misinformation}}.
\newblock {\emph{\JournalTitle{HKS Misinformation Review}}}
  \textbf{\bibinfo{volume}{1}}, \doiprefix\url{10.37016/mr-2020-033}
  (\bibinfo{year}{2020}).

\bibitem{Pierri2022}
\bibinfo{author}{Pierri, F.} \emph{et~al.}
\newblock \bibinfo{journal}{\bibinfo{title}{Online misinformation is linked to
  early {COVID}-19 vaccination hesitancy and refusal}}.
\newblock {\emph{\JournalTitle{Scientific Reports}}}
  \textbf{\bibinfo{volume}{12}}, \bibinfo{pages}{5966},
  \doiprefix\url{10.1038/s41598-022-10070-w} (\bibinfo{year}{2022}).

\bibitem{wang2023donotanswer}
\bibinfo{author}{Wang, Y.}, \bibinfo{author}{Li, H.}, \bibinfo{author}{Han,
  X.}, \bibinfo{author}{Nakov, P.} \& \bibinfo{author}{Baldwin, T.}
\newblock \bibinfo{journal}{\bibinfo{title}{Do-not-answer: A dataset for
  evaluating safeguards in llms}}.
\newblock {\emph{\JournalTitle{arXiv preprint 2308.13387}}}
  (\bibinfo{year}{2023}).
\newblock \eprint{2308.13387}.

\bibitem{reddy2023smartbook}
\bibinfo{author}{Reddy, R.~G.} \emph{et~al.}
\newblock \bibinfo{title}{Smartbook: {AI}-assisted situation report generation}
  (\bibinfo{year}{2023}).
\newblock \eprint{2303.14337}.

\bibitem{hengji_ketgsurvey_2022}
\bibinfo{author}{Yu, W.} \emph{et~al.}
\newblock \bibinfo{journal}{\bibinfo{title}{A survey of knowledge-enhanced text
  generation}}.
\newblock {\emph{\JournalTitle{ACM Comput. Surv.}}}
  \textbf{\bibinfo{volume}{54}}, \doiprefix\url{10.1145/3512467}
  (\bibinfo{year}{2022}).

\bibitem{realm2023}
\bibinfo{author}{Guu, K.}, \bibinfo{author}{Lee, K.}, \bibinfo{author}{Tung,
  Z.}, \bibinfo{author}{Pasupat, P.} \& \bibinfo{author}{Chang, M.-W.}
\newblock \bibinfo{journal}{\bibinfo{title}{Realm: retrieval-augmented language
  model pre-training}}.
\newblock {\emph{\JournalTitle{ArXiv}}}  (\bibinfo{year}{2020}).

\bibitem{filippova-2020-controlled}
\bibinfo{author}{Filippova, K.}
\newblock \bibinfo{title}{Controlled hallucinations: learning to generate
  faithfully from noisy data}.
\newblock In \emph{\bibinfo{booktitle}{Findings of the Association for
  Computational Linguistics: EMNLP 2020}}, \bibinfo{pages}{864--870},
  \doiprefix\url{10.18653/v1/2020.findings-emnlp.76}
  (\bibinfo{publisher}{Association for Computational Linguistics},
  \bibinfo{address}{Online}, \bibinfo{year}{2020}).

\bibitem{Gou2023}
\bibinfo{author}{Gou, Z.} \emph{et~al.}
\newblock \bibinfo{title}{Critic: large language models can self-correct with
  tool-interactive critiquing}.
\newblock In \emph{\bibinfo{booktitle}{arxiv}} (\bibinfo{year}{2023}).

\bibitem{Cohen2023}
\bibinfo{author}{Cohen, R.}, \bibinfo{author}{Hamri, M.},
  \bibinfo{author}{Geva, M.} \& \bibinfo{author}{Globerson, A.}
\newblock \bibinfo{title}{Lm vs lm: detecting factual errors via cross
  examination}.
\newblock In \emph{\bibinfo{booktitle}{arxiv}} (\bibinfo{year}{2023}).

\bibitem{Du2023}
\bibinfo{author}{Du, Y.}, \bibinfo{author}{Li, S.}, \bibinfo{author}{Torralba,
  A.}, \bibinfo{author}{Tenenbaum, J.~B.} \& \bibinfo{author}{Mordatch, I.}
\newblock \bibinfo{title}{Improving factuality and reasoning in language models
  through multiagent debate}.
\newblock In \emph{\bibinfo{booktitle}{arxiv}} (\bibinfo{year}{2023}).

\bibitem{Dziri2021}
\bibinfo{author}{Dziri, N.}, \bibinfo{author}{Madotto, A.},
  \bibinfo{author}{Za{\"\i}ane, O.} \& \bibinfo{author}{Bose, A.~J.}
\newblock \bibinfo{title}{Neural path hunter: reducing hallucination in
  dialogue systems via path grounding}.
\newblock In \emph{\bibinfo{booktitle}{Proceedings of the 2021 Conference on
  Empirical Methods in Natural Language Processing}},
  \bibinfo{pages}{2197--2214}, \doiprefix\url{10.18653/v1/2021.emnlp-main.168}
  (\bibinfo{publisher}{Association for Computational Linguistics},
  \bibinfo{address}{Online and Punta Cana, Dominican Republic},
  \bibinfo{year}{2021}).

\bibitem{KE2021}
\bibinfo{author}{De~Cao, N.}, \bibinfo{author}{Aziz, W.} \&
  \bibinfo{author}{Titov, I.}
\newblock \bibinfo{title}{Editing factual knowledge in language models}.
\newblock In \emph{\bibinfo{booktitle}{Proceedings of the 2021 Conference on
  Empirical Methods in Natural Language Processing}},
  \bibinfo{pages}{6491--6506}, \doiprefix\url{10.18653/v1/2021.emnlp-main.522}
  (\bibinfo{publisher}{Association for Computational Linguistics},
  \bibinfo{address}{Online and Punta Cana, Dominican Republic},
  \bibinfo{year}{2021}).

\bibitem{MEND2022}
\bibinfo{author}{Mitchell, E.}, \bibinfo{author}{Lin, C.},
  \bibinfo{author}{Bosselut, A.}, \bibinfo{author}{Finn, C.} \&
  \bibinfo{author}{Manning, C.~D.}
\newblock \bibinfo{journal}{\bibinfo{title}{Fast model editing at scale}}.
\newblock {\emph{\JournalTitle{arXiv preprint arXiv:2110.11309}}}
  (\bibinfo{year}{2021}).

\bibitem{SERAC2022}
\bibinfo{author}{Mitchell, E.}, \bibinfo{author}{Lin, C.},
  \bibinfo{author}{Bosselut, A.}, \bibinfo{author}{Manning, C.~D.} \&
  \bibinfo{author}{Finn, C.}
\newblock \bibinfo{title}{Memory-based model editing at scale}.
\newblock In \emph{\bibinfo{booktitle}{International Conference on Machine
  Learning}}, \bibinfo{pages}{15817--15831} (\bibinfo{organization}{PMLR},
  \bibinfo{year}{2022}).

\bibitem{ROME2022}
\bibinfo{author}{Meng, K.}, \bibinfo{author}{Bau, D.},
  \bibinfo{author}{Andonian, A.} \& \bibinfo{author}{Belinkov, Y.}
\newblock \bibinfo{journal}{\bibinfo{title}{Locating and editing factual
  associations in gpt}}.
\newblock {\emph{\JournalTitle{Advances in Neural Information Processing
  Systems}}} \textbf{\bibinfo{volume}{35}}, \bibinfo{pages}{17359--17372}
  (\bibinfo{year}{2022}).

\bibitem{MEMIT2023}
\bibinfo{author}{Meng, K.}, \bibinfo{author}{Sharma, A.~S.},
  \bibinfo{author}{Andonian, A.}, \bibinfo{author}{Belinkov, Y.} \&
  \bibinfo{author}{Bau, D.}
\newblock \bibinfo{journal}{\bibinfo{title}{Mass-editing memory in a
  transformer}}.
\newblock {\emph{\JournalTitle{arXiv preprint arXiv:2210.07229}}}
  (\bibinfo{year}{2022}).

\bibitem{schmidt-2019-generalization}
\bibinfo{author}{Schmidt, F.}
\newblock \bibinfo{title}{{Generalization in Generation: A closer look at
  Exposure Bias}}.
\newblock In \emph{\bibinfo{booktitle}{Proceedings of the 3rd Workshop on
  Neural Generation and Translation}}, \bibinfo{pages}{157--167},
  \doiprefix\url{10.18653/v1/D19-5616} (\bibinfo{publisher}{Association for
  Computational Linguistics}, \bibinfo{address}{Hong Kong},
  \bibinfo{year}{2019}).

\bibitem{yu2023self}
\bibinfo{author}{Yu, P.} \& \bibinfo{author}{Ji, H.}
\newblock \bibinfo{journal}{\bibinfo{title}{Self information update for large
  language models through mitigating exposure bias}}.
\newblock {\emph{\JournalTitle{arXiv preprint arXiv:2305.18582}}}
  (\bibinfo{year}{2023}).

\bibitem{zhang2020bertscore}
\bibinfo{author}{Zhang, T.}, \bibinfo{author}{Kishore, V.},
  \bibinfo{author}{Wu, F.}, \bibinfo{author}{Weinberger, K.~Q.} \&
  \bibinfo{author}{Artzi, Y.}
\newblock \bibinfo{title}{Bertscore: Evaluating text generation with {BERT}}.
\newblock In \emph{\bibinfo{booktitle}{8th International Conference on Learning
  Representations, {ICLR} 2020, Addis Ababa, Ethiopia, April 26-30, 2020}}
  (\bibinfo{publisher}{OpenReview.net}, \bibinfo{year}{2020}).

\bibitem{zhao-etal-2019-moverscore}
\bibinfo{author}{Zhao, W.} \emph{et~al.}
\newblock \bibinfo{title}{{M}over{S}core: Text generation evaluating with
  contextualized embeddings and earth mover distance}.
\newblock In \emph{\bibinfo{booktitle}{Proceedings of the 2019 Conference on
  Empirical Methods in Natural Language Processing and the 9th International
  Joint Conference on Natural Language Processing (EMNLP-IJCNLP)}},
  \bibinfo{pages}{563--578}, \doiprefix\url{10.18653/v1/D19-1053}
  (\bibinfo{publisher}{Association for Computational Linguistics},
  \bibinfo{address}{Hong Kong, China}, \bibinfo{year}{2019}).

\bibitem{oaisec}
\bibinfo{author}{OpenAI}.
\newblock \bibinfo{title}{{S}ecurity \& privacy}.
\newblock \bibinfo{howpublished}{\url{https://openai.com/security}}.
\newblock \bibinfo{note}{[Accessed 27-09-2023]}.

\bibitem{datapriv}
\bibinfo{author}{Sebastian, G.}
\newblock \bibinfo{journal}{\bibinfo{title}{Privacy and data protection in
  chatgpt and other ai chatbots: Strategies for securing user information}}.
\newblock {\emph{\JournalTitle{International Journal of Security and Privacy in
  Pervasive Computing}}} \textbf{\bibinfo{volume}{15}}, \bibinfo{pages}{1--14},
  \doiprefix\url{10.4018/IJSPPC.325475} (\bibinfo{year}{2023}).

\bibitem{europaEURLex52021PC0206}
\bibinfo{title}{{E}{U}{R}-{L}ex - 52021{P}{C}0206 - {E}{N} - {E}{U}{R}-{L}ex
  --- eur-lex.europa.eu}.
\newblock
  \bibinfo{howpublished}{\url{https://eur-lex.europa.eu/legal-content/EN/TXT/?uri=CELEX:52021PC0206}}.

\bibitem{wang2023m4}
\bibinfo{author}{Wang, Y.} \emph{et~al.}
\newblock \bibinfo{journal}{\bibinfo{title}{{{M4}: Multi-generator,
  Multi-domain, and Multi-lingual Black-Box Machine-Generated Text
  Detection}}}.
\newblock {\emph{\JournalTitle{arXiv:2305.14902}}}  (\bibinfo{year}{2023}).

\bibitem{Fung2021}
\bibinfo{author}{Fung, Y.} \emph{et~al.}
\newblock \bibinfo{title}{Infosurgeon: cross-media fine-grained information
  consistency checking for fake news detection}.
\newblock In \emph{\bibinfo{booktitle}{Proc. The Joint Conference of the 59th
  Annual Meeting of the Association for Computational Linguistics and the 11th
  International Joint Conference on Natural Language Processing (ACL-IJCNLP
  2021)}} (\bibinfo{year}{2021}).

\bibitem{misinformation2023}
\bibinfo{author}{Huang, K.-H.}, \bibinfo{author}{McKeown, K.},
  \bibinfo{author}{Nakov, P.}, \bibinfo{author}{Choi, Y.} \&
  \bibinfo{author}{Ji, H.}
\newblock \bibinfo{title}{Faking fake news for real fake news detection:
  propaganda-loaded training data generation}.
\newblock In \emph{\bibinfo{booktitle}{Proc. The 61st Annual Meeting of the
  Association for Computational Linguistics (ACL2023) Findings}}
  (\bibinfo{year}{2023}).

\bibitem{groh2023human}
\bibinfo{author}{Groh, M.} \emph{et~al.}
\newblock \bibinfo{title}{Human detection of political speech deepfakes across
  transcripts, audio, and video} (\bibinfo{year}{2023}).
\newblock \eprint{2202.12883}.

\bibitem{sadasivan2023aigenerated}
\bibinfo{author}{Sadasivan, V.~S.}, \bibinfo{author}{Kumar, A.},
  \bibinfo{author}{Balasubramanian, S.}, \bibinfo{author}{Wang, W.} \&
  \bibinfo{author}{Feizi, S.}
\newblock \bibinfo{journal}{\bibinfo{title}{Can ai-generated text be reliably
  detected?}}
\newblock {\emph{\JournalTitle{arXiv preprint arXiv:2303.11156}}}
  (\bibinfo{year}{2023}).

\bibitem{hussain2020adversarial}
\bibinfo{author}{Hussain, S.}, \bibinfo{author}{Neekhara, P.},
  \bibinfo{author}{Jere, M.}, \bibinfo{author}{Koushanfar, F.} \&
  \bibinfo{author}{McAuley, J.}
\newblock \bibinfo{title}{Adversarial deepfakes: Evaluating vulnerability of
  deepfake detectors to adversarial examples}.
\newblock In \emph{\bibinfo{booktitle}{Proceedings of the IEEE/CVF winter
  conference on applications of computer vision}}, \bibinfo{pages}{3348--3357}
  (\bibinfo{year}{2021}).

\bibitem{contentauthenticityContentAuthenticity}
\bibinfo{title}{{C}ontent {A}uthenticity {I}nitiative ---
  contentauthenticity.org}.
\newblock \bibinfo{howpublished}{\url{https://contentauthenticity.org/}}.
\newblock \bibinfo{note}{[Accessed 08-09-2023]}.

\bibitem{chatgptitaly2}
\bibinfo{title}{{ChatGPT: OpenAI reopens the platform in Italy guaranteeing
  more transparency and more rights to European users and non-users ---
  tbs-sct.canada.ca}}.
\newblock
  \bibinfo{howpublished}{\url{https://www.garanteprivacy.it/home/docweb/-/docweb-display/docweb/9881490}}.
\newblock \bibinfo{note}{[Accessed 27-09-2023]}.

\bibitem{ftcChatbotsDeepfakes}
\bibinfo{title}{{C}hatbots, deepfakes, and voice clones: {A}{I} deception for
  sale --- ftc.gov}.
\newblock
  \bibinfo{howpublished}{\url{https://www.ftc.gov/business-guidance/blog/2023/03/chatbots-deepfakes-voice-clones-ai-deception-sale}}.
\newblock \bibinfo{note}{[Accessed 08-09-2023]}.

\bibitem{canadadir}
\bibinfo{title}{{D}irective on {A}utomated {D}ecision-{M}aking ---
  tbs-sct.canada.ca}.
\newblock
  \bibinfo{howpublished}{\url{https://www.tbs-sct.canada.ca/pol/doc-eng.aspx?id=32592}}.
\newblock \bibinfo{note}{[Accessed 27-09-2023]}.

\bibitem{newsclaims2022}
\bibinfo{author}{Gangi~Reddy, R.} \emph{et~al.}
\newblock \bibinfo{title}{{N}ews{C}laims: A new benchmark for claim detection
  from news with attribute knowledge}.
\newblock In \emph{\bibinfo{booktitle}{Proceedings of the 2022 Conference on
  Empirical Methods in Natural Language Processing}},
  \bibinfo{pages}{6002--6018}, \doiprefix\url{10.18653/v1/2022.emnlp-main.403}
  (\bibinfo{publisher}{Association for Computational Linguistics},
  \bibinfo{address}{Abu Dhabi, United Arab Emirates}, \bibinfo{year}{2022}).

\bibitem{sundriyal-etal-2022-empowering}
\bibinfo{author}{Sundriyal, M.}, \bibinfo{author}{Kulkarni, A.},
  \bibinfo{author}{Pulastya, V.}, \bibinfo{author}{Akhtar, M.~S.} \&
  \bibinfo{author}{Chakraborty, T.}
\newblock \bibinfo{title}{Empowering the fact-checkers! automatic
  identification of claim spans on {T}witter}.
\newblock In \emph{\bibinfo{booktitle}{Proceedings of the 2022 Conference on
  Empirical Methods in Natural Language Processing}},
  \bibinfo{pages}{7701--7715}, \doiprefix\url{10.18653/v1/2022.emnlp-main.525}
  (\bibinfo{publisher}{Association for Computational Linguistics},
  \bibinfo{address}{Abu Dhabi, United Arab Emirates}, \bibinfo{year}{2022}).

\bibitem{sundriyal2021desyr}
\bibinfo{author}{Sundriyal, M.}, \bibinfo{author}{Singh, P.},
  \bibinfo{author}{Akhtar, M.~S.}, \bibinfo{author}{Sengupta, S.} \&
  \bibinfo{author}{Chakraborty, T.}
\newblock \bibinfo{title}{Desyr: definition and syntactic representation based
  claim detection on the web}.
\newblock In \emph{\bibinfo{booktitle}{Proceedings of the 30th ACM
  International Conference on Information \& Knowledge Management}},
  \bibinfo{pages}{1764--1773} (\bibinfo{year}{2021}).

\bibitem{Huang2022}
\bibinfo{author}{Huang, K.-H.}, \bibinfo{author}{Zhai, C.} \&
  \bibinfo{author}{Ji, H.}
\newblock \bibinfo{title}{{CONCRETE}: Improving cross-lingual fact-checking
  with cross-lingual retrieval}.
\newblock In \emph{\bibinfo{booktitle}{Proceedings of the 29th International
  Conference on Computational Linguistics}}, \bibinfo{pages}{1024--1035}
  (\bibinfo{publisher}{International Committee on Computational Linguistics},
  \bibinfo{address}{Gyeongju, Republic of Korea}, \bibinfo{year}{2022}).

\bibitem{factualerrorcorrection2023}
\bibinfo{author}{Huang, K.-H.}, \bibinfo{author}{Chan, H.~P.} \&
  \bibinfo{author}{Ji, H.}
\newblock \bibinfo{title}{Zero-shot faithful factual error correction}.
\newblock In \emph{\bibinfo{booktitle}{Proceedings of the 61st Annual Meeting
  of the Association for Computational Linguistics (Volume 1: Long Papers)}},
  \bibinfo{pages}{5660--5676}, \doiprefix\url{10.18653/v1/2023.acl-long.311}
  (\bibinfo{publisher}{Association for Computational Linguistics},
  \bibinfo{address}{Toronto, Canada}, \bibinfo{year}{2023}).

\bibitem{zhang2023stance}
\bibinfo{author}{Zhang, B.}, \bibinfo{author}{Ding, D.} \&
  \bibinfo{author}{Jing, L.}
\newblock \bibinfo{journal}{\bibinfo{title}{How would stance detection
  techniques evolve after the launch of chatgpt?}}
\newblock {\emph{\JournalTitle{arXiv preprint arXiv:2212.14548}}}
  (\bibinfo{year}{2022}).

\bibitem{KOCON2023101861}
\bibinfo{author}{Kocoń, J.} \emph{et~al.}
\newblock \bibinfo{journal}{\bibinfo{title}{{ChatGPT}: Jack of all trades,
  master of none}}.
\newblock {\emph{\JournalTitle{Information Fusion}}}
  \textbf{\bibinfo{volume}{99}}, \bibinfo{pages}{101861},
  \doiprefix\url{https://doi.org/10.1016/j.inffus.2023.101861}
  (\bibinfo{year}{2023}).

\bibitem{itnextRememberingConversations}
\bibinfo{author}{Shankar, A.}
\newblock \bibinfo{title}{{R}emembering {C}onversations: {B}uilding {C}hatbots
  with {S}hort and {L}ong-{T}erm {M}emory on {A}{W}{S} --- itnext.io}.
\newblock
  \bibinfo{howpublished}{\url{https://itnext.io/remembering-conversations-building-chatbots-with-short-and-long-term-memory-on-aws-c1361c130046}}.
\newblock \bibinfo{note}{[Accessed 28-09-2023]}.

\bibitem{ferrara2023genai}
\bibinfo{author}{Ferrara, E.}
\newblock \bibinfo{journal}{\bibinfo{title}{{{GenAI} Against Humanity:
  Nefarious Applications of Generative Artificial Intelligence and Large
  Language Models}}}.
\newblock {\emph{\JournalTitle{arXiv preprint arXiv:2310.00737}}}
  (\bibinfo{year}{2023}).

\bibitem{nationalacademiesLoginNational}
\bibinfo{title}{{L}ogin | {T}he {N}ational {A}cademies {P}ress ---
  nap.nationalacademies.org}.
\newblock
  \bibinfo{howpublished}{\url{https://nap.nationalacademies.org/download/24623}}.
\newblock \bibinfo{note}{[Accessed 08-09-2023]}.

\bibitem{bbcChatGPTBanned}
\bibinfo{author}{McCallum, S.}
\newblock \bibinfo{title}{{C}hat{G}{P}{T} banned in {I}taly over privacy
  concerns}.
\newblock
  \bibinfo{howpublished}{\url{https://www.bbc.com/news/technology-65139406}}.
\newblock \bibinfo{note}{[Accessed 08-09-2023]}.

\bibitem{universityworldnewsUniversityPrinciples}
\bibinfo{title}{{N}ew {U}{K} university principles promote {A}{I} literacy and
  integrity --- universityworldnews.com}.
\newblock
  \bibinfo{howpublished}{\url{https://www.universityworldnews.com/post.php?story=20230704155107330}}.
\newblock \bibinfo{note}{[Accessed 08-09-2023]}.

\bibitem{genAIeducationUK}
\bibinfo{author}{{Department for Education}}.
\newblock \bibinfo{title}{Generative artificial intelligence in education
  departmental statement}.
\newblock
  \bibinfo{howpublished}{\url{https://assets.publishing.service.gov.uk/government/uploads/system/uploads/attachment_data/file/1146540/Generative_artificial_intelligence_in_education_.pdf}}.
\newblock \bibinfo{note}{[Accessed 08-09-2023]}.

\bibitem{nvidiaRightTrack}
\bibinfo{author}{Cohen, J.}
\newblock \bibinfo{title}{{R}ight on {T}rack: {N}{V}{I}{D}{I}{A}
  {O}pen-{S}ource {S}oftware {H}elps {D}evelopers {A}dd {G}uardrails to {A}{I}
  {C}hatbots --- blogs.nvidia.com}.
\newblock
  \bibinfo{howpublished}{\url{https://blogs.nvidia.com/blog/2023/04/25/ai-chatbot-guardrails-nemo/}}.
\newblock \bibinfo{note}{[Accessed 08-09-2023]}.

\bibitem{Chen2023-fj}
\bibinfo{author}{Chen, A.} \& \bibinfo{author}{Chen, D.~O.}
\newblock \bibinfo{journal}{\bibinfo{title}{Accuracy of chatbots in citing
  journal articles}}.
\newblock {\emph{\JournalTitle{JAMA Netw Open}}} \textbf{\bibinfo{volume}{6}},
  \bibinfo{pages}{e2327647} (\bibinfo{year}{2023}).

\bibitem{microsoftIntroducingMicrosoft}
\bibinfo{author}{Spataro, J.}
\newblock \bibinfo{title}{{I}ntroducing {M}icrosoft 365 {C}opilot – your
  copilot for work - {T}he {O}fficial {M}icrosoft {B}log ---
  blogs.microsoft.com}.
\newblock
  \bibinfo{howpublished}{\url{https://blogs.microsoft.com/blog/2023/03/16/introducing-microsoft-365-copilot-your-copilot-for-work/}}.
\newblock \bibinfo{note}{[Accessed 08-09-2023]}.

\bibitem{Pacheco2021Coordinated}
\bibinfo{author}{Pacheco, D.} \emph{et~al.}
\newblock \bibinfo{title}{Uncovering coordinated networks on social media:
  Methods and case studies}.
\newblock In \emph{\bibinfo{booktitle}{Proc. International AAAI Conference on
  Web and Social Media (ICWSM)}}, vol.~\bibinfo{volume}{15},
  \bibinfo{pages}{455--466}, \doiprefix\url{10.1609/icwsm.v15i1.18075}
  (\bibinfo{year}{2021}).

\end{thebibliography}

\end{document}